\documentstyle[twoside,cl,lingmacros]{article}  
\issue{22}{2}{1996}
\title{Limited Attention and Discourse Structure \\
\begin{small} {\tt cmp-lg/9512003} \end{small} }

\author{Marilyn A. Walker
\thanks{ATT Laboratories, 600 Mountain Ave.,
Murray Hill, N.J. 07974, USA, {\tt walker@research.att.com}}}

\runningtitle{Limited Attention and Discourse Structure}
\runningauthor{Marilyn A. Walker}

\begin{document}           
\maketitle                 
\bibliographystyle{fullname}  



\section{Hierarchical versus Linear Recency}
\label{intro-sec}

In computational theories of discourse, there are at least three
processes presumed to operate under a {\sc limited attention
constraint} of some type: (1) ellipsis interpretation; (2) pronominal
anaphora interpretation; and (3) inference of discourse relations
between representations A and B of utterances in a discourse, e.g. B
{\sc motivates} A.  In each case, the
interpretation of the current element B of a discourse depends on the
accessibility of another earlier element A.  According to the {\sc
limited attention constraint} only a limited number of candidates need
to be considered in the processing of B, e.g. only a limited number of
entities in the discourse model are potential cospecifiers for a
pronoun.

The limited attention constraint has been defined by some researchers
by {\sc linear recency}: a representation of an utterance A is {\sc
linearly recent} for a representation of an utterance B if A is
linearly adjacent to B.  Using linear recency as a model of the
limited attention constraint would mean that an antecedent for an
anaphor is determined by a linear backward search of the text, or of a
discourse model representation of the text \cite{CS79} {\it inter
alia}.

In contrast, other work has formulated the limited attention
constraint in terms of {\sc hierarchical recency}
\cite{GS86,Hobbs85a,MannThompson87} {\it inter alia}.  A
representation of an utterance A is {\sc hierarchically recent} for a
representation of an utterance B if A is adjacent to B in the tree
structure of the discourse. Of all theories based on hierarchical
recency, only Grosz and Sidner's theory of discourse structure
provides an operationalization of hierarchical recency in terms of
their {\sc stack} model of attentional state
\cite{Sidner79,Grosz77,GS86}. Thus,  below, the relationship
between limited attention and hierarchical recency will be discussed
in terms of their stack model, but the discussion should also apply to
claims about the role of hierarchical recency in other work.

In the remainder of this squib, I will argue that the limited
attention constraint must account for three types of evidence: (1) the
occurrence of {\sc informationally redundant utterances} in naturally
occurring dialogues \cite{Walker93c}; (2) the infelicity of discourses
that depend on accessing discourse entities that are not linearly
recent; and (3) experiments that show that humans have limited
attentional capacity \cite{Miller56,Baddeley86}.

\section{Evidence for Limited Attention from Anaphoric Processing}

\begin{figure}[htb]
\begin{small}
\begin{center}
\begin{tabular}{|l||l|}
\hline &\\
 Dialogue A  & Dialogue B \\
\hline \hline &\\
(4) C: Ok Harry, I'm have a problem & (4) C: Ok Harry, I'm have a problem\\
       that uh my - with today's economy  &   that uh my - with today's
economy\\
       {\it my daughter is working}, & {\it my daughter is working}\\
(5) H: I missed your name. & (5) H: I missed your name.\\
(6) C: Hank. &  (6) C: Hank\\
& (6.2) H: Is that H A N K? \\
& (6.3) C: Yes. \\
(7) H: Go ahead Hank &  (7) H: Go ahead Hank\\
(8a) C: {\it as well as her uh husband}. & (8a) C: {\it as well as her uh
husband}. \\
(8b) They have a child. &  (8b) They have a child.\\
(8c) and they bring the child to us  & (8c) and they bring the child to us\\
every day for babysitting. & every day for babysitting.\\
\hline
\end{tabular}
\end{center}
\end{small}
\caption{Dialogue A does not include the extra utterances shown
in Dialogue B. Dialogue B is identical to A except for utterances 6.2, 6.3.}
\label{examp-ab-fig}
\end{figure}

In figure \ref{examp-ab-fig}, Dialogue A, hierarchical recency
supports the interpretation of the proforms in utterance (8a) from a
radio talk show for financial advice \cite{PHW82}. In utterance A-5, H
interrupts C's narrative to ask for his name, but in A-8, C continues
as though A-4 had just been said.  Utterance A-8a realizes the
proposition {\it My daughter's husband is working as well}, but this
realization depends on both an anaphoric referent and an anaphoric
property.

According to the stack model, since utterances A-5 $\ldots$ A-7 are
part of an embedded segment, A-4 is {\sc hierarchically recent} when
A-8 is interpreted.   A new focus space is pushed
on the stack during the processing of dialogue A when the intention of
utterance 5 is recognized. Since utterance 7 clearly indicates
completion of the interrupting segment, the focus space for the
interruption in 5 to 7 is popped from the stack after utterance 7,
leaving the focus space for utterances 1 to 4 on the top of the stack.
This focus space supports the interpretation of the proforms in A-8a.

However, consider the variation of dialogue A in dialogue B in figure
\ref{examp-ab-fig}.  Here, the segment between B-5 $\ldots$
B-7 is also an embedded segment.  Utterance B-7 indicates completion of
the embedded segment and signals a pop.  So, by the stack model, this
segment is handled by the same focus stack popping mechanism as we saw
for dialogue A.

However, in dialogue B, utterance 8a is more difficult, if not
impossible, to interpret.  This is surprising because utterance B-4 is
hierarchically recent for B-8a, just as it is in dialogue A. The
interruption in dialogue B is but a slightly longer version of that in
dialogue A. Inasmuch as the stack model is a precise formulation of
hierarchical recency, it does not predict the infelicity of dialogue
B. The problem arises partly because the stack model includes no
constraints related to the length, depth, or the amount of processing
required for an embedded segment.  Thus, these types of extended
embedded segments suggest that the limited attention constraint must
be sensitive to some aspect of linear recency.

\section{Evidence for Limited Attention from Informational Redundancy}

Additional evidence for the influence of linear recency arises from
the analysis of {\sc informationally redundant} utterances (IRUs) in
naturally-occurring discourse \cite{Walker93c}.\footnote{A subclass of
Attention IRUs, Open-Segment IRUs, is discussed here.} IRUs realize
propositions already established as mutually believed in the
discourse. IRUs have {\sc antecedents} in the discourse, which are
those utterances that originally established the propositions realized
by the IRU as mutually believed.  Consider excerpt C from the
financial advice corpus.  Here E has been telling H about how her
money is invested, and then poses a question in C-3.  IRUs in the
examples below are capitalized and their antecedents are italicized.


\begin{small}
\eenumsentence[C]
{\item[]
( 3) E: And I was wondering -- should I continue on with the certificates or \\
( 4) H: Well it's difficult to tell because we're so far away from
    any of them -- but I would suggest this -- if {\it all of these are 6
    month certificates and I presume they are} \\
( 5) E: {\it Yes} \\
( 6) H: {\it Then I would like to see you start spreading some of that
    money around} \\
( 7) E: uh huh \\
( 8) H: Now in addition, how old are you? \\
        . \\
        (discussion and advice about retirement investments)\\
        . \\
(21) E: uh huh and  \\
(22a) H: But as far as the certificates are concerned,  \\
(22b) I'D LIKE THEM SPREAD OUT A LITTLE BIT  - \\
(22c) THEY'RE ALL 6 MONTH CERTIFICATES  \\
(23) E: Yes \\
(24) H: And I don't like putting all my eggs in one basket...
\label{cert-examp}
}
\end{small}

The utterances in 22b and 22c realize propositions previously
established as mutually believed, so they are IRUs.\footnote{The
antecedents are in utterances 4, 5 and 6: H asserted the content of
22b to E in 6.  E indicated understanding and implicated acceptance of
this assertion in 7 \cite{Walker92a}, and E confirmed the truth of the
content of 22c for H in 5.} The cue word {\it but} in utterance 22a
indicates a push, a new intention \cite{GS86}.  The phrase {\it as far
as the certificates are concerned} indicates that this new intention
is subordinate to the previous discussion of the certificates.  Thus,
utterance 22a, {\it but as far as the certificates are concerned}, has
the effect that the focus space related to the discussion of
retirement investments, corresponding to utterances 8 to 21, is
popped from the stack.

This means that the focus space representation of the intention for
utterances 4 to 7 are on the top of the stack after C-22a, when 22b
and 22c are processed.  Therefore there are two reasons why it is
surprising that H restates the content of utterances 4, 5 and 6 in 22b
and 22c: (1) The propositions realized by 22b and 22c are already
mutually believed; and (2) These mutual beliefs should be salient by
virtue of being on top of the stack. If they are salient by virtue of
being on top of the stack, they should be accessible for processes
such as content-based inferences or the inference of discourse
relations. If they must be accessible for these inferences to take
place, as I will argue below, their reintroduction suggests that in
fact they are not accessible.  Many similar examples of IRUs are found
in the corpus \cite{Walker93c}. Thus, these types of IRUs show that
hierarchical recency, as realized by the stack model, does not predict
when information is accessible.

\section{The Cache Model of Attentional State}
\label{cache-sec}

The evidence above suggests the need for a model of attentional state
in discourse that reflects the limited attentional capacity of human
processing.  Here, I propose an alternate model to the stack model,
which I will call the {\sc cache model}, and discuss the evidence for
this model. In section \ref{discuss-sec}, I compare a number of
dimensions of the cache and stack models.

The notion of a cache in combination with main memory, as is standard
in computational architectures, is a good basis for a computational
model of human attentional capacity in processing discourse.  All
conversants in a dialogue have their own cache and some conversational
processes are devoted to keeping these caches synchronized.

The cache model consists of: (1) basic mechanisms and architectural
properties; (2) assumptions about processing; (3) specification of
which mechanism is applied at which point.  The {\sc cache} represents
working memory and {\sc main memory} represents long-term memory. The
cache is a limited capacity, almost instantaneously accessible, memory
store.  The exact specification of this capacity must be determined by
future work, but previous research suggests a limit of 2 or 3
sentences, or approximately 7 propositions \cite{Kintsch88,Miller56}.
Main memory is larger than the cache, but is slower to access
\cite{Baddeley86,Kintsch88}.

There are three operations involving the cache and main memory. Items
in the cache can be preferentially {\sc retained} and items in main
memory can be {\sc retrieved} to the cache. Items in the cache can
also be {\sc stored} to main memory.

When new items are retrieved from main memory to the cache, or enter
the cache directly due to events in the world, other items may be {\sc
displaced}, because the cache has limited capacity.  Displaced items
are stored in main memory. The determination of which items to
displace is handled by a cache replacement policy.  The specification
of the cache replacement policy is left open, however, replacing items
that haven't been recently used, with the exception of those items
that are preferentially retained, is a good working assumption, as
shown by previous work on linear recency.\footnote{Obviously, linear
recency is simply an approximation to what is in the cache.  If
something has been recently discussed, it was recently in the cache,
and thus is is more likely to still be in the cache than other items.
However, linear recency ignores the effects of retention and
retrieval.}

The cache model includes specific assumptions about processing.
Discourse processes execute on elements that are in the cache.  All of
the premises for an inference must be simultaneously in the cache for
the inference to be made \cite{McKoonRatcliff92,Walker93c}. If a
discourse relation is to be inferred between two separate segments, a
representation of both segments must be simultaneously in the cache
\cite{FHM90,Walker93c}. The cospecifier of an anaphor must
be in the cache for automatic interpretation, or be strategically
retrieved to the cache in order to interpret the anaphor
\cite{GMR92}.  Thus what is contained in the cache at any one
time is a {\sc working set} consisting of discourse entities such as
entities, properties and relations that are currently being used for
some process.

Two factors determine when cache operations are applied: (1) the
speaker's intentions and the hearer's recognition of intention; (2)
expectations about what will be discussed.

The cache model maintains the distinction between intentional
structure and attentional state first proposed by Grosz and Sidner
(1986). This distinction is critical.  Just as a cache can be used for
processing the references and operations of a hierarchically
structured program, so can a cache be used to model attentional state
when discourse intentions are hierarchically structured. The
intentions of a conversant and the recognition of the other's
intentions determine what is {\sc retrieved} from main memory and what
is preferentially {\sc retained} in the cache.

When conversants start working towards the achievement of a new
intention, that intention may utilize information that was already in
the cache. If so, that information will be preferentially retained in
the cache because it is being used.  Whenever the new intention
requires information that is not currently in the cache, that
information must be retrieved from main memory. Thus the process of
initiating the achievement of the new intention has the result that
some, and perhaps all, of the items currently in the cache are
replaced with items having to do with the new intention.

When conversants return to a prior intention, information relevant to
that intention must be retrieved from main memory if it has not been
retained in the cache.

When an intention is completed, it is not necessary to strategically
retain information relevant to the completed segment in the cache. It
does not necessarily mean that there is an automatic retrieval of
information related to other intentions.  However, automatic retrieval
processes can be triggered by associations between information being
currently discussed and information stored in main memory
\cite{GMR92}. These processes make items salient that have not been explicitly
mentioned.

{\sc expectations} about what will be discussed also determine
operations on the cache. Expectations can arise from shared knowledge
about the task, and from the prior discourse
\cite{Grosz77,Malt84}. Expectations can arise from
interruptions when the nature of the interruption makes it obvious
that there will be a return to the interrupted segment. When the
pursuit of an intention is momentarily interrupted, as in dialogue A,
the conversants attempt to retain the relevant material in the cache
during the interruption.

\section{Evaluating Critical Evidence: comparing the cache with the stack}

\label{discuss-sec}

In this section, I wish to examine evidence for the cache model, look
at further predictions of the model,and then discuss evidence relevant
to both stack and cache models in order to draw direct comparisons
between them.  First, I contrast the mechanisms of the models with
respect to certain discourse processes.

\begin{itemize}
\item New intention subordinate to current intention: (1) Stack pushes
new focus space; (2) Cache retrieves entities related to new intention

\item Intention completed: (1) Stack pops focus space for intention
from stack, entities in focus space are no longer accessible; (2)
Cache doesn't retain entities for completed intention, but they remain
accessible until displaced

\item New intention subordinate to prior intention: (1)  Stack pops
 focus spaces for intervening segments, focus space for prior
intention accessible after pop; (2) Cache retrieves entities related
to prior intention from main memory to cache, unless retained in the
cache

\item Informationally redundant utterances: (1) Stack predicts no role
for IRUs when they are represented in focus space on top of stack,
because information should be immediately available; (2) Cache
predicts that IRUs reinstantiate or refresh known information in the
cache

\item Returning from interruption: (1)  In the stack model, the length
 and depth of the interruption and the processing required is
irrelevant; (2) In the cache model, the length of the interruption or
the processing required predicts retrievals from main memory
\end{itemize}

First, consider the differences in the treatment of interruptions.
The state of the stack when returning from an interruption is
identical for interruptions of various lengths and depths of
embedding. In the cache model, an interruption may give rise to an
expectation of a return to a prior intention, and each participant may
attempt to retain information relevant to pursuing that intention in
their  cache. However, it may not be possible to retain the relevant
material in the cache.  In dialogue B, the interruption is too long
and the working set for the interruption uses all of the cache. When
this happens, the relevant material is displaced to main memory. On
returning after an interruption, the conversants must initiate a cued
retrieval of beliefs and intentions.  This will require some
processing effort, yielding the prediction that there will be a short
period of time in which the cache does not have the necessary
information. This would mean that the processing of incoming
information would be slower until all of the required information is
in the cache.\footnote{This could predict the observed occurrence of
disfluencies at segment boundaries \cite{PassonneauLitman94}.}
The ease with which the conversants can return to a previous
discussion will then rely on the retrievability of the required
information from main memory, and this in turn depends on what is
stored in main memory and the type of cue provided by the speaker as
to what to retrieve. For example, if processing involves the surface
form of the utterance, as it might dialogue B, we can explain the
clear-cut infelicity by the fact that surface forms are not normally
stored in main memory
\cite{Sachs67}.

Next, consider the differences between the models with respect to the
function of IRUs.  In dialogue C, a version of the dialogue without
the IRUs is possible but is harder to interpret.  Consider dialogue C
without 22b, 22c and 23, i.e. replace 22a to 24 with {\it But as far
as the certificates are concerned, I don't like all my eggs in one
basket.} Interpreting this alternate version requires the same
inference, namely that having all your investments in six month
certificates constitutes the negatively evaluated condition of having
all your eggs in one basket. However the inference requires more
effort to process.

The stack model doesn't predict a function for the IRUs.  However,
according to the cache model, IRUs make information accessible that is
not accessible by virtue of hierarchical recency, so that processes of
content-based inferences, inference of discourse relations, and
interpretation of anaphors can take place with less effort.  Thus, one
prediction of the cache model is that a natural way to make the
anaphoric forms in dialogue B more easily interpretable is to
re-realize the relevant proposition with an IRU, as in 8a':{\it My
problem is that my daughter is working, as well as her uh husband. }

The IRU may function this way since: (1) the IRU reinstantiates the
necessary information in the cache; or (2) the IRU is a retrieval cue
for retrieval of information to the cache. Here reinstantiation is
certainly sufficient, but in general these cases cannot be
distinguished from corpus analysis. It should be possible to test
psychologically using reaction time methods, whether and under what
conditions IRUs function to simply reinstantiate an entity in the
cache, and when they serve as retrieval cues.

Next, consider the differences in status of the entities in completed
discourse segments. In the stack model, focus spaces for segments that
have been closed are popped from the stack and entities in those focus
spaces are not accessible.  In the cache model, ``popping'' only
occurs via displacement. Thus even when a segment is clearly closed,
if a new topic has not been initiated, the popped entities should
still be available. Some support for the cache model predictions about
popped entities is that (1) rules proposed for deaccenting noun
phrases treat popped entities as accessible \cite{DavisHirschberg88};
and (2) Rules for referring expressions in argumentative texts treat
the conclusions of popped sisters as salient \cite{Huang94}. Stronger
evidence would be the reaction times to the mention of entities in a
closed segment, after it is clear that a new segment has been
initiated, but before the topic of that new segment has initiated a
retrieval to, and hence displacement from, the cache.

It should also be possible to test whether entities that are in the
focus spaces on the stack, according to the stack model, are more
accessible than entities that have been popped off the stack.  In the
cache model, the entities in these focus spaces would not have a
privileged attentional status, unless of course they had been
refreshed in the cache by being realized implicitly or explicitly in
the intervening discussion.

Finally, consider one of the most studied predictions of the stack
model: cases where a pronoun has an antecedent in a prior focus space.
These cases have been called {\sc return pops} or {\sc focus pops}
\cite{Grosz77,Sidner79,Reichman85,Fox87,PassonneauLitman94}.  In the
stack model, any of the focus spaces on the stack can be returned to,
and the antecedent for a pronoun can be in any of these focus spaces.
As a potential alternative to the stack model, the cache model appears
to be unable to handle return pops since a previous state of the cache
can't be popped to. Since return pops are a primary motivation for the
stack model, I will re-examine all of the naturally-occurring return
pops that I was able to find in the literature. There are 21 of them.
While it would be premature to draw final conclusions from such a
small sample size, I will argue that the data supports the conclusion
that return pops are {\bf cued retrieval from main memory} and that
the cues reflect the context of the pop \cite{RatcliffMcKoon88}. Thus,
return pops are not problematic for the cache model.

In the cache model, there are at least three possibilities for how the
context is created so that pronouns in {\sc return pops} can be
interpreted: (1) The pronoun alone functions as a retrieval cue
\cite{GMR92}; or (2) The content of the first utterance in a return
indicates what information to retrieve from main memory to the cache,
which implies that the interpretation of the pronoun is delayed; (3)
The shared knowledge of the conversants creates expectations that
determines what is in the cache, e.g.  shared knowledge of the task
structure.

Let us consider the first possibility.  The view that pronouns must be
able to function as retrieval cues is contrary to the view that
pronouns indicate entities that are currently salient
\cite{Prince81}.  However, there are certain cases where a
pronoun alone is a good retrieval cue, such as when only one referent
of a particular gender or number has been discussed in the
conversation. If competing antecedents are those that match the
gender and number of the pronoun \cite{Fox87}, then only 11 of the 21
return pops found in the literature have competing antecedents.

Thus, the numbers suggest that in about half the cases we could expect
the pronoun to function as an adequate retrieval cue based on gender
and number alone.  In fact, Sidner proposed that return pops might
always have this property in her {\sc stacked focus constraint}:
``Since anaphors may co-specify the focus or a potential focus, an
anaphor which is intended to co-specify a stacked focus must not be
acceptable as co-specifying either the focus or potential focus. If,
for example, the focus is a noun phrase which can be mentioned with an
{\it it} anaphor, then {\it it} cannot be used to co-specify with a
stacked focus.'' \cite{Sidner79}, p. 88,89.

In addition, the representation of the anaphor should include
selectional restrictions from the verb's subcategorization frame as
retrieval cues \cite{Dieugenio90}.  Of the 11 tokens with
competing antecedents, 5 tokens have no competing antecedents if
selectional restrictions are also applied.  For example, in the
dialogues about the construction of a pump from
\cite{Grosz77}, only some entities can be bolted, loosened, or made to
work.  Only 4 pronouns of the 21 return pops have competing referents
if a selectional constraint can arise from the dialogue, e.g.  if only
one of the male discourse entities under discussion has been riding a
bike, then the verb {\it rode} serves as a cue for retrieving that
entity \cite{PassonneauLitman94}. Thus in 17 cases, an adequate
retrieval cue is constructed from processing the pronoun and the
matrix verb \cite{Dieugenio90}.

The second hypothesis is that the content of the return utterance
indicates what information to retrieve from main memory to the
cache. The occurrence of IRUs as in dialogue C is one way of doing
this.  IRUs at the locus of a return can: (1) reinstantiate required
information in the cache so that no retrieval is necessary; (2)
function as excellent retrieval cues for information from main memory.
An examination of the data shows that IRUs occur in 6 of the 21 return
pops.  IRUs in combination with selectional restrictions leave only 2
cases of pronouns in return pops with competing antecedents.

In the remaining 2 cases, the competing antecedent is not and was
never prominent in the discourse, i.e. it was never the discourse
center, suggesting that it may never compete with the other
cospecifier.

It should be possible to test how long it takes to resolve anaphors in
return pops and under what conditions it can be done, considering the
data presented here on competing referents, IRUs, explicit closing,
and selectional restrictions.  A probe just after a pronoun in a
return pop and before the verb could determine whether the pronoun
alone is an adequate retrieval cue, or whether selectional information
from the verb is required or simply speeds processing.

Finally, it should be possible to test whether pronouns in return pops
are accented, which signals to the hearer that the most recent
antecedent is not the correct one \cite{Cahn91}.

To conclude, the analysis presented here suggests many hypotheses that
could be empirically tested, which the currently available evidence
does not enable us to resolve.

\section{Discussion and Conclusion}

This squib has discussed the role of limited attention in a
computational model of discourse processing. The cache model was
proposed as a computational implemention of human working memory and
operations on attentional state are formulated as operations on a
cache.  Just as a cache can be used for processing the references and
operations of a hierarchically structured program, so can a cache be
used to model attentional state when discourse intentions are
hierarchically structured.

The store and retrieve operations of the cache model casts discourse
processing as a {\bf gradient} phenomenon, predicting that the
contents of the cache will change gradually, and that change requires
processing effort.  The notion of processing effort for retrieval
operations on main memory makes predictions that can be experimentally
tested. In the meantime, the notion of increased processing effort in
the cache model explains the occurrence of a class of {\sc
informationally redundant} utterances in discourse, as well as cases
of infelicitous discourses constructed as variations on naturally
occurring ones, while remaining consistent with evidence on human
limited attentional capacity.  Finally, the cache model appears to
handle the class of ``return pops'' which prima facie should be
problematic for the model.

\starttwocolumn

\begin{acknowledgments}

I'd like to thank Aravind Joshi, Ellen Prince, Mark Liberman, Karen
Sparck Jones, Bonnie Webber, Scott Weinstein, Susan Brennan, Janet
Cahn, Mitch Marcus, Cindie McLemore, Owen Rambow, Candy Sidner, Ellen
Germain, Megan Moser, Becky Passonneau, Pam Jordan, Jennifer Arnold,
and Steve Whittaker for extended discussion of the issues in this
paper. Thanks also to the two anonymous reviewers.
\end{acknowledgments}

\end{document}